\relax
\documentclass[letterpaper]{article} 
\usepackage{aaai21}  
\usepackage{times}  
\usepackage{helvet} 
\usepackage{courier}  
\usepackage[hyphens]{url}  
\usepackage{graphicx} 
\usepackage{natbib}  
\usepackage{caption} 
\usepackage{hyperref}
\hypersetup{
	colorlinks=true,
	linkcolor=black,     
	urlcolor=blue,
	citecolor=black,
}
\usepackage{pgfplots}
\pgfplotsset{compat=1.16}
\usetikzlibrary{external}
\usepackage{subfig}         

\usepackage{tabularx,booktabs}
\newcolumntype{Y}{>{\centering\arraybackslash}X}
\usepackage{multirow}       

\usepackage{amsfonts}	
\usepackage{amsmath}	
\usepackage{amssymb}	
\usepackage{amsthm}
\theoremstyle{definition}

\usepackage{bm}

\let\conjugate\overline

\usepackage[acronym]{glossaries}

\newacronym{cvnn}{CVNN}{Complex-Valued Neural Network}
\newacronym{rvnn}{RVNN}{Real-Valued Neural Network}
\newacronym{cv-mlp}{CV-MLP}{Complex-Valued Multi-Layer Perceptron}
\newacronym{autodiff}{autodiff}{automatic differentiation}
\newacronym{api}{API}{Application Programming Interface}
\newacronym{ux}{UX}{User eXperience}
\newacronym{sgd}{SGD}{Stochastic Gradient Descent}
\newacronym{cn}{CN}{Complex Normal}
\newacronym{mlp}{MLP}{MultiLayer Perceptron}
\newacronym{cnn}{CNN}{Convolutional Neural Networks}
\newacronym{relu}{ReLU}{Rectified Linear Unit}
\newacronym{tanh}{tanh}{hyperbolic tangent}
\newacronym{1hl}{1HL}{one hidden layer}
\newacronym{2hl}{2HL}{two hidden layers}

\title{Complex-Valued vs. Real-Valued Neural Networks for Classification Perspectives: An Example on Non-Circular Data}

\author{
    J. A. Barrachina,\textsuperscript{\rm 1}\textsuperscript{\rm 2}\thanks{The authors would like to thank the D\'el\'egation G\'en\'erale de l'Armement (DGA) for funding.}
    C. Ren,\textsuperscript{\rm 1}
    C. Morisseau,\textsuperscript{\rm 2}
    G. Vieillard,\textsuperscript{\rm 2}
    J.-P. Ovarlez\textsuperscript{\rm 1}\textsuperscript{\rm 2} \\
}

\affiliations{%
    \textsuperscript{\rm 1} SONDRA, CentraleSup\'elec, Universit\'e Paris-Saclay, rue Joliot Curie, 91192 Gif-sur-Yvette, France \\%
    \textsuperscript{\rm 2} DEMR, ONERA, Universit\'e Paris-Saclay, Chemin de la Hunière, 91120 Palaiseau, France \\
}

\begin{document}
\maketitle

\begin{abstract}
    The contributions of this paper are twofold.
	First, we show the potential interest of \acrfull{cvnn} on classification tasks for complex-valued datasets.
	To highlight this assertion, we investigate an example of complex-valued data in which the real and imaginary parts are statistically dependent through the property of non-circularity. 
    In this context, the performance of fully connected feed-forward \acrshort{cvnn}s is compared against a real-valued equivalent model.
    The results show that \acrshort{cvnn} performs better for a wide variety of architectures and data structures. 
	\acrshort{cvnn} accuracy presents a statistically higher mean and median and lower variance than \acrfull{rvnn}. 
	Furthermore, if no regularization technique is used, \acrshort{cvnn} exhibits lower overfitting.
	The second contribution is the release of a \href{https://github.com/NEGU93/cvnn}{Python library} \cite{negu2019cvnn} using \textit{Tensorflow} as back-end that enables the implementation and training of \acrshort{cvnn}s in the hopes of motivating further research on this area.
\end{abstract}

\section{Introduction}

In the Machine Learning community, most Neural Networks are developed for processing real-valued features (voice signals, RGB images, videos, etc.). 
The signal processing community, however, has a higher interest in developing theory and techniques over complex fields. Indeed, complex-valued signals are encountered in a large variety of applications such as biomedical sciences, physics, communications and radar. 
All these fields use signal processing tools \cite{SS2010},  which are usually based on complex filtering operations (Discrete Fourier Transform, Wavelet Transform, Wiener Filtering, Matched Filter, etc.). 
\acrshort{cvnn}s appear as a natural choice to process and to learn from these complex-valued features since the operation performed at each layer of \acrshort{cvnn}s can be interpreted as complex filtering. Notably, \acrshort{cvnn}s are more adapted to extract phase information \cite{hirose2012generalization}, which could be helpful, \textit{e.g.}, for retrieving Doppler frequency in radar signals, classifying polarimetric SAR data \cite{hansch2009classification, zhao2019contrastive}, etc.  

Although \acrshort{cvnn} has been proven to achieve better results than its real-valued counterpart for particular structures of complex-valued data \cite{hirose2012generalization, hirose2012complex, hirose2009complex, hansch2009classification}, the difficulties in implementing \acrshort{cvnn} models in practice have slowed down the field from growing further \cite{monning2018evaluation}. In this paper, we address this issue by 
providing a \textit{Python} library to deal with the implementation of \acrshort{cvnn} models.
The code also offers statistical indicators, e.g., loss and accuracy box plots \cite{mcgill1978variations, chambers2018graphical}, to compare the performance of \acrshort{cvnn} models with its real-valued counterpart.

There are no rules in practice to prioritize \acrshort{cvnn} over \acrshort{rvnn} for complex-valued datasets. Performance highly depend on the characteristics of the complex-valued dataset. 
Indeed, merely taking real data as input does not profit from using \acrshort{cvnn} \cite{monning2018evaluation}.
We propose in this paper to analyze the influence of the non-circular statistical property on the performance of both \acrshort{cvnn} and \acrshort{rvnn} networks. We show that particular structures of such complex data, as, for example, phase information or statistical correlation between real and imaginary parts, can notably benefit from using \acrshort{cvnn} compared to its real-valued equivalent.
Under this context, \acrshort{cvnn} is potentially an attractive network to obtain better classification performance on complex datasets.

The paper is organized as follows: In Section \ref{sec:math-back}, we discuss the mathematical background of \acrshort{cvnn} to then continue with the circularity property for a random variable. 
Section \ref{sec:experim-setup} discusses the feed-forward network architecture and the programming code used for these experiments. Section \ref{sec:experim-results} illustrates the comparison of statistical performance obtained for each network. 
In particular, the sensitivity of \acrshort{cvnn} and \acrshort{rvnn} results are evaluated either by changing the dataset characteristics or the network hyper-parameters. 

Although \acrshort{cvnn} is an acronym that involves numerous complex-valued neural network architectures, in this work, we will always be referring to fully connected feed-forward ones. This convention was chosen to be coherent with the existing bibliography \cite{hirose2012generalization, hirose2012complex, hirose2009complex, hansch2009classification, amin2011, hirose2013complex}.

\section{Mathematical Background} \label{sec:math-back}

A natural way to build \acrshort{cvnn} consists in extending \acrshort{rvnn} for handling complex-valued neurons. The latter implies that the weights for connecting neurons and the hidden activation functions should be complex-valued. 
In contrast, the loss function remains real-valued to minimize an empirical risk during the learning process. Despite the architectural change for handling complex-valued inputs, the main challenge of \acrshort{cvnn} is the way to train such neural networks.

\subsection{Wirtinger derivation}

When training \acrshort{cvnn}s, the weights are updated using a complex gradient, so the back-propagation operation becomes complex-valued.
In complex analysis, \textit{Liouville's theorem} states that every bounded holomorphic function is constant \cite[pp.70-71]{FR1997}, 
implying that the loss and activation functions should be either constant or unbounded.

\textit{Wirtinger calculus} \cite{WIRTINGER_CALCULUS} generalizes the notion of complex derivative for \textit{non-holomorphic} functions.
It states that given a complex function $f(z)$ of a complex variable $z = x + j\,y \in \mathbb{C}, (x, y) \in \mathbb{R}^2$, the partial derivatives with respect to $z$ and $\conjugate{z}$ respectively are:
\begin{equation}\label{eq:wirtinger-calculus}
	\begin{aligned}
	    \frac{\partial f}{\partial z} &\triangleq \frac{1}{2}\, \left( \frac{\partial f}{\partial x} - j \, \frac{\partial f}{\partial y} \right),
	    \frac{\partial f}{\partial \conjugate{z}} &\triangleq \frac{1}{2}\, \left( \frac{\partial f}{\partial x} + j \, \frac{\partial f}{\partial y} \right) \, .
	\end{aligned}
\end{equation}

\textit{Wirtinger calculus} enables one to work with \textit{non-holomorphic} functions, providing an alternative method for computing the gradient. 
Following references \cite{amin2011} and \cite{amin_learning_2013}, the complex gradient is then defined as:
\begin{equation}
	\nabla_z f = 2 \, \frac{\partial f}{\partial \conjugate{z}}\; .
	\label{eq:complex-gradient-def}
\end{equation}

Also, for any real-valued loss function $\mathcal{L}: \mathbb{C} \to \mathbb{R}$, the complex derivative of the composition of $\mathcal{L}$ with any complex function $g:\mathbb{C} \to \mathbb{C}$ with $g(z)=r(z)+j \, s(z)$ is given by the following so-called chain rule:
\begin{equation}
	\frac{\partial \mathcal{L} \circ g}{\partial \conjugate{z}} = \frac{\partial \mathcal{L}}{\partial r} \, \frac{\partial r}{\partial \conjugate{z}} + \frac{\partial \mathcal{L}}{\partial s} \, \frac{\partial s}{\partial \conjugate{z}}\; .
	\label{eq:gradient-real-out}
\end{equation}

Using equations \ref{eq:complex-gradient-def} and \ref{eq:gradient-real-out} it is possible to train \acrshort{cvnn}.

\subsection{Circularity} \label{sec:circularity}

The importance of circularity for \acrshort{cvnn}s has already been mentioned in \cite{hirose2012complex,hirose2013complex}. Let us denote the vector $\mathbf{u} \triangleq [X, Y]^T$ as the real random vector built by stacking the real and imaginary parts of a complex random variable $Z = X + j \, Y$. The probability density function (pdf) of $Z$ can be identified with the pdf of $\mathbf{u}$. 
The \textit{variance} of $Z$ is defined by: 
\begin{equation}
    \sigma_Z^2 \triangleq \mathbb{E}\left[\left|Z-\mathbb{E}[Z]\right|^2 \right] = \sigma_X^2 + \sigma_Y^2,
\end{equation}
where $\sigma_X^2$ and $\sigma_Y^2$ are respectively the \textit{variance} of $X$ and $Y$. The latter
does not bring information about the \textit{covariance}: 
\begin{equation}
\sigma_{XY} \triangleq \mathbb{E}\left[\left(X-\mathbb{E}[X]\right) \,  \left(Y-\mathbb{E}[Y]\right) \right],
\end{equation}
but this information can be retrieved thanks to the  \textit{pseudo-variance} \cite{ollila2008circularity, picinbono1996secondordercomplex}: 
\begin{equation}
\tau_{Z} \triangleq \mathbb{E}\left[\left(Z-\mathbb{E}[Z]\right)^2 \right] = \sigma_X^2 - \sigma_Y^2 + 2 \, j\, \sigma_{XY}.
\end{equation}
The circularity quotient $\varrho_Z$ is then: $\varrho_Z = \tau_Z/\sigma_Z^2$. If  $Z$ has a  vanishing \textit{pseudo-variance}, $\tau_{Z}=0$, or equivalently, $\varrho_Z=0$, it is said to be second-order circular. 
The correlation coefficient is defined as
\begin{equation}
    \rho =\frac{\sigma_{XY}}{\sigma_X \sigma_Y}. 
\end{equation}

In this work, complex non-circular random datasets are generated and classified, with two non-exclusive possible sources of non-circularity: $X$ and $Y$ have unequal variances or $X$ and $Y$ are correlated. 

\section{Proposed Neural Networks} \label{sec:experim-setup}

\subsection{Library} \label{sec:library}

Although it has been proven several times that \acrshort{cvnn} achieves better generalization than \acrshort{rvnn} \cite{hirose2012generalization}, the latter has been often favored over the former due to difficulties in the implementation \cite{monning2018evaluation}. Indeed, the two most popular 
\href{https://www.python.org/}{Python} 
libraries for developing deep neural networks, 
\href{https://pytorch.org/}{Pytorch} 
and 
\href{https://www.tensorflow.org/}{Tensorflow}
 do not fully support the creation of complex-valued models.
Therefore, one of the contributions of this work is to provide a library, published in conjunction with this paper \cite{negu2019cvnn}, that enables the full implementation of fully-connected feed-forward \acrshort{cvnn}s using \textit{Tensorflow} as back-end. 

Even though \textit{Tensorflow} does not fully support the implementation of a \acrshort{cvnn}, it has one significant benefit: 
It enables the use of complex-valued data types for the \acrfull{autodiff} algorithm \cite{hoffmannHitchhikerGuideAutomatic2016} to calculate the complex gradients.
\textit{Tensorflow}'s \acrshort{autodiff} implementation calculates the gradient for $f:\mathbb{C} \to \mathbb{C}$ as:
\begin{equation}
	\nabla_z f = \conjugate{\left. \frac{\partial f}{\partial z} +  \frac{\partial \conjugate{f}}{\partial z}\right.} \; .
	\label{eq:tf-complex-gradient-def}
\end{equation}

It is straightforward to prove that \eqref{eq:complex-gradient-def}  and \eqref{eq:tf-complex-gradient-def} are equivalent for a real-valued loss function $f:\mathbb{C} \to \mathbb{R}$ \cite{kreutz2009complex}. 

Libraries to develop \acrshort{cvnn}s do exist, the most important one of them being probably the \href{https://github.com/ChihebTrabelsi/deep_complex_networks}{code} published in \cite{trabelsi2017deep}. However, we have decided not to use this library since the latter uses \textit{Theano} as back-end, which is no longer \href{https://groups.google.com/g/theano-users/c/7Poq8BZutbY/m/rNCIfvAEAwAJ}{maintained}. Another \href{https://github.com/JesperDramsch/keras-complex}{code} was published on \href{https://github.com/}{GitHub} that uses \textit{Tensorflow} and \textit{Keras} to implement \acrshort{cvnn} \cite{dramsch2019complex}. However, as \textit{Keras} does not support complex-valued numbers, the published code simulates complex operations always using real-valued datatypes. Therefore, the user has to transform its complex-valued data to a real-valued equivalent before using this library.

The library used to generate the results of this paper was tested through several well known real-valued applications. Probably the most relevant test was the adaptation of \textit{Tensorflow}'s online code example \href{https://www.tensorflow.org/datasets/keras_example}{Training a neural network on MNIST with Keras} since the model architecture is very similar to the one used in this paper.
The network architecture has been coded with our library. Statistically, the results obtained with our network were equivalent to the one given by \textit{Tensorflow}'s example. They can be found in the \textit{GitHub} repository inside the \textit{examples} subfolder.

For more information about the library published in this paper, please refer to the \href{https://complex-valued-neural-networks.readthedocs.io/en/latest/index.html}{documentation}.

\subsection{Model Architecture}\label{sec:model-arch}

\subsubsection{Activation function}

One of the essential characteristics of \acrshort{cvnn} is its activation functions, which should be non-linear and complex-valued. Additionally, the latter is desired to be piece-wise smooth to compute the gradient. The extension to complex field offers many possibilities to design an activation function. Among them, two main types are proposed by extending real-valued activation functions \cite{kuroe2003activation}:
\begin{itemize}
	\item Type-A: $g_A(f) = g_R\left(\mathrm{Re}{(f)}\right) + j \, g_I(\mathrm{Im}{(f)})$,
	\item Type-B: $g_B(f) = g_r(|f|) \, e^{j \, g_\phi(\phi(f))}$,
\end{itemize}
where $f: \mathbb{C} \to \mathbb{C}$ is a complex function and $g_{R},g_I,g_r,g_{\phi}$ are all real-valued  functions\footnote{Although not with the same notation, these two types of complex-valued activation functions are also discussed in Section 3.3 of \cite{hirose2012complex}}. 
The most popular activation functions, sigmoid, \acrfull{tanh} and \acrfull{relu}, are extensible using Type-A or Type-B approach.

Normally, $g_\phi$ is left as a linear mapping \cite{kuroe2003activation, hirose2012complex}. Under this condition, using \acrshort{relu} activation function for $\sigma_r$ has a limited interest since the latter makes $g_B$ converge to a linear function, limiting Type-B \acrshort{relu} usage.
Nevertheless, \acrshort{relu} has increased in popularity over the others as it has proved to learn several times faster than equivalents with saturating neurons \cite{krizhevsky2012imagenet}. 
Consequently, we will adopt, in Section \ref{sec:experim-results}, the Type-A \acrshort{relu}, which is a non-linear function, as \acrshort{cvnn} hidden layers activation functions.

The image domain of the output layer depends on the set of data labels. For classification tasks, real-valued integers or binary numbers are frequently used to label each class. Therefore the output layer's activation function should be real-valued. For this reason, we use \textit{softmax} (normalized exponential) as the output activation function, which maps the magnitude of complex-valued input to $[0;1]$, so the image domain is homogeneous to a probability.

\subsubsection{Number of layers}



Even though the tendency is to make the models as deep as possible for \acrfull{cnn}, this is not the case for fully-connected feed-forward neural networks, also known as \acrfull{mlp}. For these models, \acrfull{1hl} is usually sufficient for the vast majority of problems \cite{hornikMultilayerFeedforwardNetworks1989, stinchcombe1989universal}. Although some authors may argue that \acrfull{2hl} may be better than one \cite{thomasTwoHiddenLayers2017}, all authors seem to agree that
there is currently no theoretical reason to use a \acrshort{mlp} with more than two hidden layers \cite[p.~158]{heaton2008introduction}

References \cite{heaton2008introduction} and \cite{kulkarni2015artificial} recommend the neurons of the hidden layer to be between the size of the input layer and the output layer. 
Therefore, two models will be used as default in section \ref{sec:experim-results}, one with a single hidden layer of size 64 and one with two hidden layers of shape 100 and 40 for the first and second hidden layers respectively.
In order to prevent the models from overfitting, dropout regularization technique \cite{srivastavaDropoutSimpleWay2014} is used on \acrshort{1hl} and \acrshort{2hl} hidden layers. Both \acrshort{cvnn} and \acrshort{rvnn} are trained with a dropout factor of 0.5. 


\subsubsection{Equivalent RVNN} \label{sec:equiv-rvnn}

To define an equivalent \acrshort{rvnn}, the strategy used in \cite{hirose2012complex} is adopted, separating the input $z$ into two real values $(x, y)$ where $z = x + j\, y$, giving the network a double amount of inputs. The same is done for the number of neurons in each hidden layer.
Although this strategy keeps the same amount of features in hidden layers, it provides a higher capacity for the \acrshort{rvnn} with respect to the number of real-valued training parameters \cite{monning2018evaluation}. 

\subsubsection{Loss function and optimizer}

Mean square error and Cross-Entropy loss functions are mostly used for \acrshort{rvnn} to solve regression and classification problems, respectively. 
The loss remains the same for \acrshort{cvnn} since the training phase still requires the minimization over a real-valued loss function.
We currently limit our optimizer to the well-known standard \acrfull{sgd}. The default learning rate used in this work is $0.01$ as being \textit{Tensorflow}'s default (v2.1) for its 
\acrshort{sgd} \href{https://www.tensorflow.org/api_docs/python/tf/keras/optimizers/SGD}{implementation}. 

\subsubsection{Weights initialization}

For weights initialization, Glorot uniform (also known as Xavier uniform) \cite{glorot2010understanding} is used, and all biases start at zero as those are \textit{Tensorflow}'s current (v2.1) default initialization methods for 
\href{https://www.tensorflow.org/api_docs/python/tf/keras/layers/Dense}{dense layers}. 
Glorot initialization generates weight values according to the uniform distribution in  \cite[eq.16]{glorot2010understanding} where its boundaries depend on both input and output sizes of the initialized layer. 
Note that \acrshort{cvnn} has halved the number of input and output neurons compared to an equivalent \acrshort{rvnn}, meaning the initial values of the real and the imaginary part of its weights are generated using a slightly larger uniform distribution boundary comparing to the real case. Therefore, \acrshort{cvnn} might be penalized by starting with a higher initial loss value before any training.

\section{Experimental Results}\label{sec:experim-results}

\subsection{Dataset setup} \label{sec:dataset}

As mentioned previously, to respect the equivalence between \acrshort{rvnn} and \acrshort{cvnn}, we use in the following input vectors of size 128 (resp. 256) for \acrshort{cvnn}s (resp. \acrshort{rvnn}).  Each element of the feature vector is generated according to a non-circular \acrlong{cn} distribution $\mathcal{CN}(0, \sigma_Z^2, \tau_Z)$. Two sources of non-circularity could occur in practice: $\sigma_X \neq \sigma_Y$ and/or  $\rho \neq 0$, or equivalently $\tau_Z \neq 0$. Therefore, we propose to evaluate the classification performance of \acrshort{cvnn} and \acrshort{rvnn} for three types of datasets presented in Table \ref{tab:datasets}.

\begin{table}[ht]
	\centering
	\begin{tabularx}{0.45\textwidth}{ c *{6}{Y} }
	    \toprule
	    \multicolumn{1}{c}{} &
	    \multicolumn{2}{c}{Data A} & \multicolumn{2}{c}{Data B} & \multicolumn{2}{c}{Data C} \\
	    \cmidrule(lr){2-3} \cmidrule(lr){4-5} \cmidrule(lr){6-7}
	    Class   & 1 & 2 & 1 & 2 & 1 & 2 \\
	    \midrule
	    $\rho$ & $0.3$ & $-0.3$ & $0$ & $0$ & $0.3$ & $-0.3$  \\
        $\sigma_X^2$ & $1$ & $1$ & $1$ & $2$ & $1$ & $2$  \\
        $\sigma_Y^2$ & $1$ & $1$ & $2$ & $1$ & $2$ & $1$ \\
        $\varrho_Z$ & $j0.3$ & $-j0.3$ & $-\frac{1}{3}$ & $\frac{1}{3}$ & $\frac{j-0.6}{3}$ & $\frac{0.6-j}{3}$ \\
        \bottomrule
    \end{tabularx}
	\caption{Dataset characteristics}	
	\label{tab:datasets}
\end{table}

\begin{figure}[ht]
    \centering
    \input{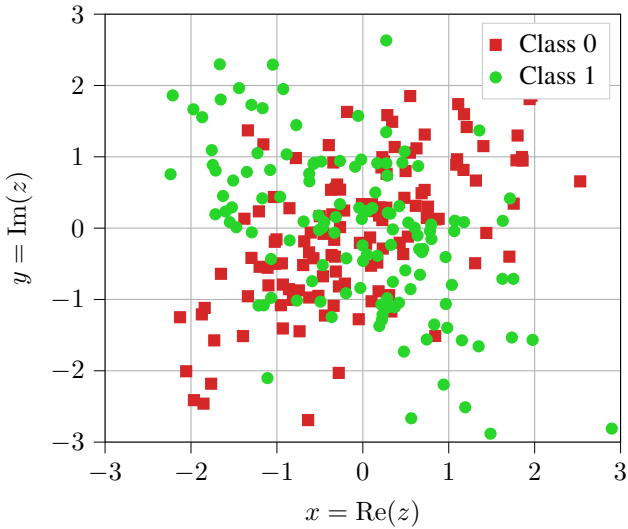}
    \caption{Dataset A example}
 	\label{fig:dataset-example}
\end{figure}

Figure \ref{fig:dataset-example} shows an example of two classes for dataset A with a feature size of 128. It is possible to see that most features are coincident even if they are from different classes, meaning that several points will yield no information or even confuse the classification algorithm. 

It is important to note that the distinction between classes is entirely contained in the relationship between the real and imaginary parts. This means that removing, for example, the imaginary part of the dataset will result in both classes being statistically identical, and therefore, rendering the classification impossible. 

To evaluate the difficulty of classifying this dataset, a Maximum Likelihood Estimation of $\tau_Z$ was implemented with the \textit{prior} knowledge of the underlying Gaussian distributions used to generate the dataset. The data are then classified using a threshold on the estimate of $\tau_Z$. The accuracy of this classifier gives an upper bound of the optimal accuracy.
For a low correlation coefficient, for example $\rho = 0.1$, this parametric classifier only achieves around $85\%$ accuracy. 


\subsection{1HL and 2HL baseline results}


To ensure that the models do not fall short of data, 10000 samples of each class were generated using $80\%$ for the train set and the remaining $20\%$ for testing. Accuracy and loss of both \acrshort{cvnn} and \acrshort{rvnn}, defined previously in section \ref{sec:model-arch}, are statistically evaluated over 30 Monte-Carlo trials. Each trial contained 300 epochs with a batch size of 100.

\begin{figure}[ht]
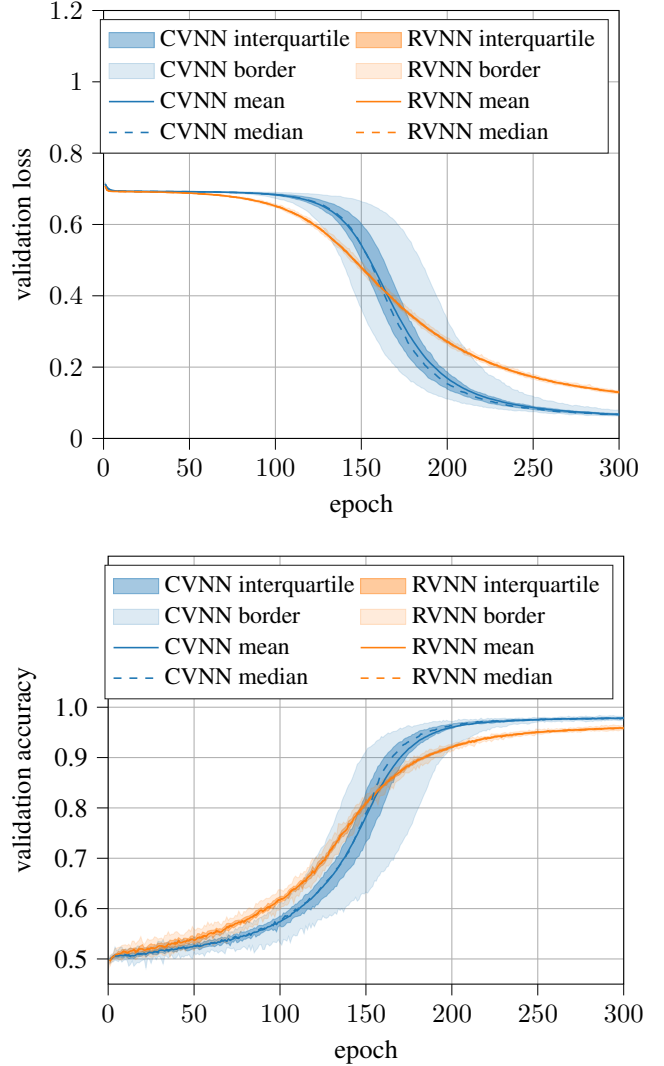

    \centering
    \subfloat{\input{dropout_loss.tikz}}\\
    \subfloat{\input{dropout_accuracy.tikz}}\\
    \caption{Test loss and accuracy for \acrshort{2hl} \acrshort{cvnn} and \acrshort{rvnn} with a dropout of 50\%}
 	\label{fig:dropout-loss-accuracy}
\end{figure}

Only the \acrshort{2hl} case is illustrated in Fig.~\ref{fig:dropout-loss-accuracy} as their results are more favorable to the \acrshort{rvnn} model. 
Although \acrshort{rvnn} starts having a higher accuracy, after enough epochs \acrshort{cvnn} surpasses and achieves higher performance.
Both losses behave well without significant indication of overfitting.
Additionally, the test accuracy of \acrshort{cvnn} trials stays above $97.30\%$ whereas \acrshort{rvnn} stays under $96.30\%$. The test accuracy of \acrshort{rvnn} is lower than the \acrshort{cvnn}.

Table \ref{tab:results} summarizes the test accuracy of \acrshort{1hl} and \acrshort{2hl} models for all three different datasets. 
The median error is computed as $1.57 \, \mathrm{IQR}/\sqrt{n}$ \cite{mcgill1978variations}, where $\mathrm{IQR}$ is the Inter-Quartile Range, and $n$ is the number of trials. 
According to \cite{chambers2018graphical}, using this error definition, if median values do not overlap there is a $95$\% confidence that their values differ.

\begin{table*}[h]
	\centering
	\begin{tabularx}{\textwidth}{ c r c c c c  c c }
	    \toprule
	    \multicolumn{2}{c}{}
	    & \multicolumn{2}{c}{Data A}
	    & \multicolumn{2}{c}{Data B}
	    & \multicolumn{2}{c}{Data C} \\
	    \cmidrule(lr){3-4} \cmidrule(lr){5-6} \cmidrule(lr){7-8}
	    \multicolumn{2}{c}{} & \acrshort{cvnn} & \acrshort{rvnn}
	    & \acrshort{cvnn} & \acrshort{rvnn}
	    & \acrshort{cvnn} & \acrshort{rvnn} \\
	    \midrule
	    \multirow{4}{*}{\acrshort{1hl}} 
	    & median 
	    &  $96.16 \pm 0.11$ & $94.48 \pm 0.06$ 
	    &  $97.39 \pm 0.07$ & $96.65 \pm 0.06$
	    &  $99.67 \pm 0.04$ & $99.48 \pm 0.03$ \\
	    & mean 
	    &  $96.20 \pm 0.04$ & $94.51 \pm 0.04$ 
	    &  $97.67 \pm 0.02$ & $96.66 \pm 0.04$
	    &  $96.66 \pm 0.01$ & $99.47 \pm 0.02$ \\
	    & IQR 
	    & $96.06-96.43$ & $94.38-94.59$ 
	    & $97.31-97.54$ & $96.57-96.78$  
	    & $99.61-99.73$ & $99.43-99.52$ \\
	    & full range 
	    & $95.65-96.60$ & $94.02-95.03$ 
	    & $97.03-97.80$ & $96.25-97.07$ 
	    & $99.50-99.77$ & $99.27-99.67$ \\
	    \midrule
	    \multirow{4}{*}{\acrshort{2hl}} 
	    & median 
	    &  $97.83 \pm 0.08$ & $95.82 \pm 0.08$      
	    &  $98.89 \pm 0.05$ & $97.83 \pm 0.05$      
	    &  $99.90 \pm 0.02$ & $99.87 \pm 0.01$  \\
	    & mean 
	    &  $97.81 \pm 0.04$ & $95.86 \pm 0.04$ 
	    &  $98.88 \pm 0.02$ & $97.82 \pm 0.03$
	    &  $99.90 \pm 0.01$ & $99.86 \pm 0.01$ \\
	    & IQR 
	    & $97.70-97.97$ & $95.71-95.97$ 
	    & $98.77-98.94$ & $97.75-97.90$ 
	    & $99.87-99.92$ & $99.84-99.87$ \\
	    & full range 
	    & $97.35-98.22$ & $95.53-96.30$  
	    & $98.65-99.05$ & $97.62-98.08$
	    & $99.85-99.98$ & $99.77-99.92$ \\
        \bottomrule
	\end{tabularx}
	\caption{Test accuracy results (\%)}
	\label{tab:results}
\end{table*}

Because the results are skewed, there is some difference between mean and median accuracy, as the mean is less robust to outliers. For this reason, the median would be a better measure of central tendency in this paper's simulations.
For the complex-valued model, the outliers tend to be the bad cases, whereas, for the real model, they are the good cases. This can be verified by the mean being lower than the median for \acrshort{cvnn} and higher than the median for \acrshort{rvnn}. 

For dataset B \acrshort{cvnn} still proves to be superior to \acrshort{rvnn}. For \acrshort{1hl}, \acrshort{cvnn} achieves very high accuracy with a median of over $97.39$\%.
Dataset C presents almost the same results as dataset A with some improvement for both architectures.


In general, both \acrshort{rvnn} and \acrshort{cvnn} performed better with \acrshort{2hl} than with \acrshort{1hl}. 

\subsection{Case without dropout}

\begin{figure}[ht]
	\centering
    \input{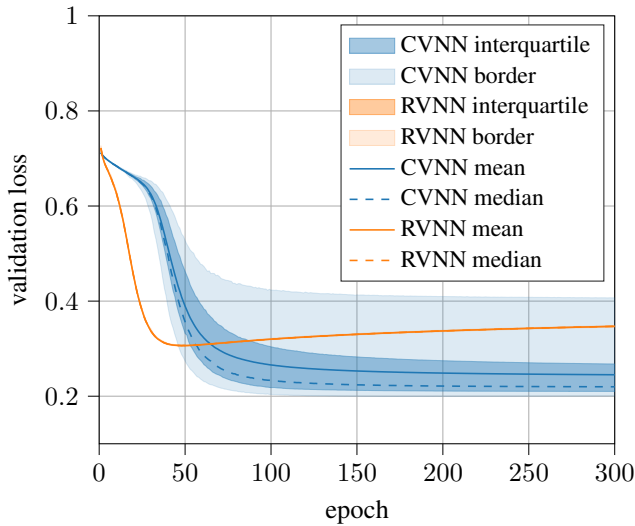}
	\caption{Test loss for \acrshort{2hl} \acrshort{cvnn} and \acrshort{rvnn} without dropout}
	\label{fig:no-dropout-loss-accuracy}
\end{figure}

The simulations were re-done with no dropout to test the models tendency to overfit. \acrshort{rvnn} presented very high overfitting, as can be seen in figure \ref{fig:no-dropout-loss-accuracy}. However, \acrshort{cvnn} presented higher variance.

The impact was so high that the test accuracy of \acrshort{rvnn} drop from $97.8\%$ to $90.15\%$ for \acrshort{2hl}, whereas for \acrshort{cvnn} it dropped less than $4\%$ reaching a test median accuracy of $93.48\%$ and mean test accuracy of $92.81$ for the case without dropout.
The totality of the simulations mentioned in this report were also done without dropout. In general, dropout had very little impact on \acrshort{cvnn}s performance and a huge amelioration for \acrshort{rvnn}. However, it is worth mentioning that although in average \acrshort{cvnn} outperformed \acrshort{rvnn} when no dropout was used, it presented many outliers with a very low accuracy.
In conclusion, \acrshort{cvnn} has a very low tendency to over-fit, whereas for \acrshort{rvnn} is decisive to use dropout or at least a regularization technique.

\subsection{Phase and amplitude}

Since the phase information could be relevant for classifying these datasets, polar-\acrshort{rvnn}, is defined where the inputs are the amplitude and phase of data. However, this method had a very high overfitting even with higher values of dropout. Regardless of the many models simulated we failed to obtain a case where polar-\acrshort{rvnn} outperformed conventional \acrshort{rvnn} so using real and imaginary part seams to be better than amplitude and phase.




\subsection{Parameter sensibility study} \label{sec:param}

In this section, a swipe through several model architectures and hyper-parameters is done to assert that the results obtained are independent of specific parameters. These simulations are done for both \acrshort{1hl} and \acrshort{2hl} networks. 

Other sensibility studies were done but are not presented in this work due to paper size limitations. They concern learning rate, dataset size, feature vector size and multi-classes for all combinations of \acrshort{1hl} and \acrshort{2hl} with and without dropout.

\subsubsection{Correlation coefficient}

\begin{figure}[ht]
	\centering
    \input{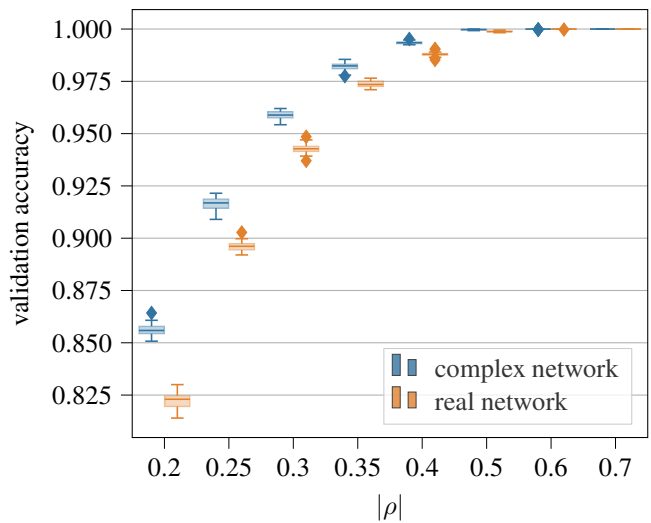}
	\caption{Test accuracy box plot for different values of correlation coefficient $\rho$ for \acrshort{1hl} model with dropout}
	\label{fig:merge-box-plot}
\end{figure}

Several correlation coefficients have been tested for \acrshort{1hl} and \acrshort{2hl} models. Figure \ref{fig:merge-box-plot} shows the accuracy of \acrshort{1hl} models tested on datasets similar to data A (see table \ref{tab:datasets}), in which the correlation coefficient varies from 0.2 to 0.7. As $|\rho|$ rises, \acrshort{cvnn} merits become evident. 
When $|\rho|$ is close to one, then the link between real and imaginary parts is strengthened, which facilitates the classification of the data for both models.

\section{Conclusion}

In this paper, we provided a end-to-end tool for the implementation of \acrshort{cvnn}s to fully use the complex-valued characteristics of the data. It also allows a more straightforward comparison with real equivalent networks in order to motivate further analysis and use of \acrshort{cvnn} \cite{negu2019cvnn}. 
Moreover we showed that \acrshort{cvnn}s stand as attractive networks to obtain higher performances than conventional \acrshort{rvnn}s on complex-valued datasets. 
The latter point was illustrated by several examples of non-circular complex-valued data which cover a large amount of data types that can be encountered in signal processing and radar fields. 
All statistical indicators showed that \acrshort{cvnn} clearly outperforms \acrshort{rvnn} showing higher accuracy and less overfitting, regardless the model architecture and hyper-parameters.
Conversely, \acrshort{rvnn} do starts with a higher accuracy on the first epochs.
Further analysis involving sensitivities to learning rates, feature vector size, multi-class dataset will be investigated to assert the generalisation of our results.

\newpage
\bibliography{biblio}

\end{document}